\documentclass[letterpaper, 10 pt, conference]{ieeeconf}

\IEEEoverridecommandlockouts     
\usepackage{amsmath,amssymb,amsfonts}
\usepackage{algorithmic}
\usepackage{graphicx}

\usepackage{multirow}
\usepackage{booktabs}

\newcommand{\magat}
{\texttt{\textsc{MAGAT}}}

\newcommand{\dmn}
{\texttt{\textsc{D2M2N}}}

\newcommand{\oracle}
{\texttt{\textsc{Oracle}}}

\title{\LARGE\bf D2M2N: Decentralized Differentiable Memory-Enabled\\ Mapping and Navigation for Multiple Robots}

\author{
  Md Ishat-E-Rabban and
  Pratap Tokekar\thanks{This work is supported by the National Science Foundation under Grant No. 1943368 and ONR under grant number N00014-18-1-2829. The authors are with Department of Computer Science, University of Maryland College Park
 \texttt{\{ier,tokekar\}@umd.edu}}
}

\begin{document}
\maketitle
\thispagestyle{empty}
\pagestyle{empty}


\begin{abstract}

Recently, a number of learning-based models have been proposed for multi-robot navigation. However, these models lack memory and only rely on the current observations of the robot to plan their actions. They are unable to leverage past observations to plan better paths, especially in complex environments. In this work, we propose a fully differentiable and decentralized memory-enabled architecture for multi-robot navigation and mapping called \dmn. \dmn\, maintains a compact representation of the environment to remember past observations and uses  Value Iteration Network for complex navigation. We conduct extensive experiments to show that \dmn\, significantly outperforms the state-of-the-art model in complex mapping and navigation task. 


\end{abstract}


\section{Introduction}
\label{intro}

Learning-based methods have recently been demonstrated to yield effective robot control policies for tasks such as navigation~\cite{neuralmap, hierarchy, prmrl}, exploration~\cite{complexenv, exploration1,shi2021communication}, flocking~\cite{sadler2021loop, flocking1, flocking2}, and coverage~\cite{coverage1,rabban2021failure,rabban2019mvfs}. These methods have shown their capabilities to offload the online computational burden into an offline learning procedure, which allows agents to act effectively based on the learned knowledge. These works provide multiple advantages over classical centralized algorithms and other distributed algorithms including faster planning and generalizability to unknown environments~\cite{li2020gnnpp}. 

In this paper, we focus on the multi-robot navigation problem in a structured environment. At each time-step, each robot makes a partial observation of the surrounding environment and chooses an action to reach its goal position. Two robots can communicate if they are within the communication range of each other. Several learning-based planners have been developed to solve this problem in an end-to-end manner. Most recent architectures~\cite{li2020gnnpp, li2021gatpp} employ a Convolutional Neural Network (CNN) to encode the robot observations, variants of Graph Neural Networks (GNN) to communicate messages, and a Multi-Layer Perceptron (MLP) to select the actions. However, these architectures only take into account the current observations, and hence cannot leverage memory of what the team has observed in the past. Furthermore, a simple MLP for selecting the action is not expressive enough to solve the navigation task in a complex occupancy map where the optimal path to the goal is significantly longer than the straight-line distance from the goal. Consequently these models exhibit poor performance in the case of complex environments, e.g., a maze.

To this end, we present a new architecture that: (i) equips each robot with individual memory, which stores a compact representation of its belief of the occupancy status of the environment; and (ii) uses a Value Iteration Network (VIN)~\cite{tamar2016vin} as the action selector module instead of an MLP. The memory works as an input to the VIN. Consequently, these two changes work in conjunction with each other.

\begin{figure}[]
\centering
\includegraphics[width=1.0\linewidth]{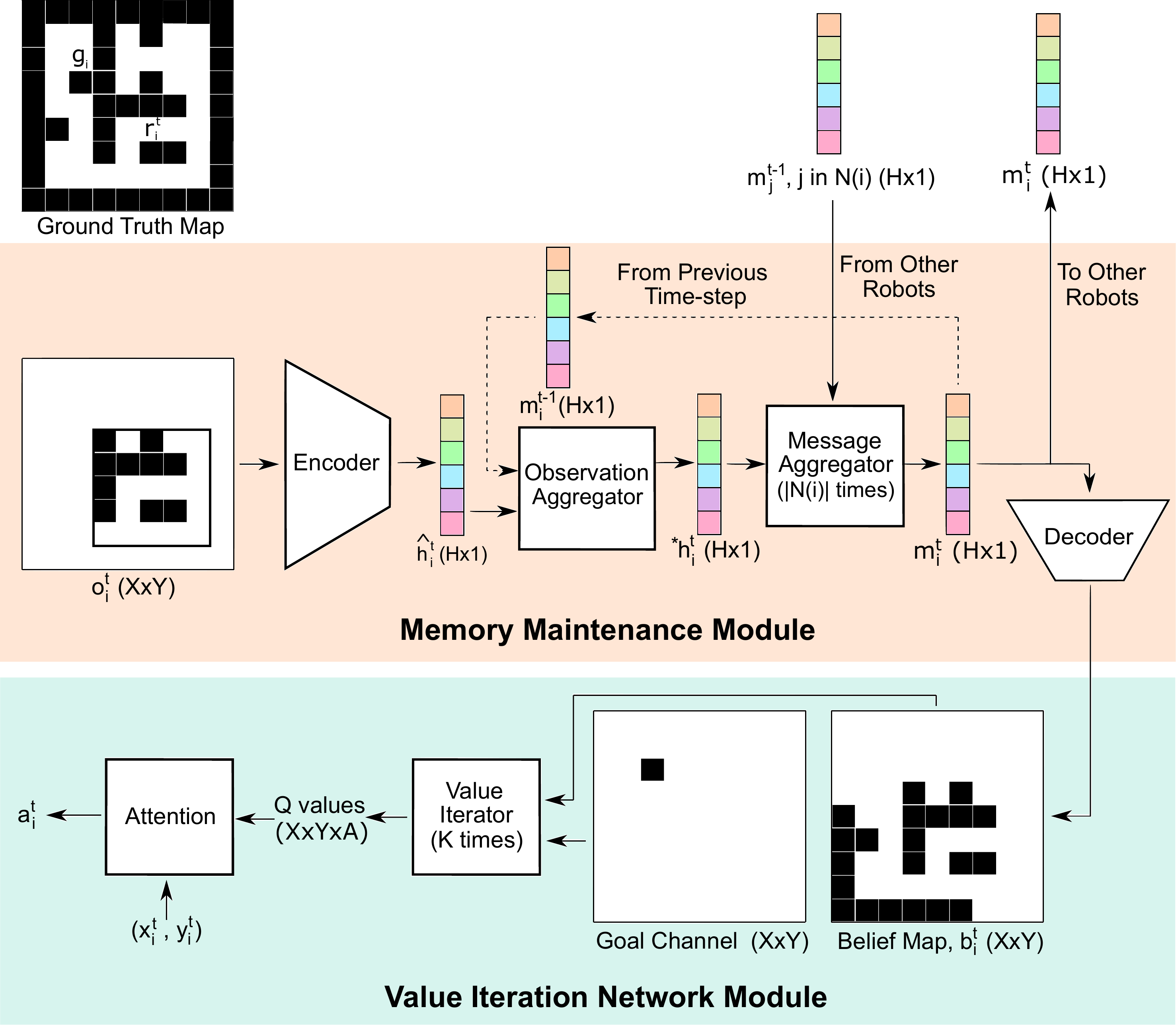}
\caption{Proposed Architecture of \dmn{} takes as input local observations and maintains a compact embedding $m_i^t$ of the map. This embedding is updated over time using new local observations and embeddings received from neighboring robots. The planner (VIN) module uses the decoded embedding to select optimal actions for the robot.}
\label{fig_arch}
\end{figure}

Our proposed architecture (Figure~\ref{fig_arch}) maintains a compressed embedding for each robot, which represents the robot's belief about the occupancy map of the environment. Two robots communicate their embeddings with each other when they are within the communication range. Exchanging embeddings, instead of the full map, helps reduce the communication overhead. The embeddings get updated with each observations made, and with each message received from a neighboring robot. Given the occupancy grid and the destination as input, the planner imitates the value iteration process~\cite{bellman1957dp} to select an action. The memory and planning modules can be trained end-to-end or separately in a supervised manner, which makes the architecture differentiable. We call this architecture \textit{Decentralized Differentiable Memory-enabled
Mapping and Navigation} (\dmn).


We conduct extensive experimentation to evaluate the performance of \dmn{} and compare it with the state-of-the-art architecture for multi-robot navigation, \magat, proposed by Li et al.~\cite{li2021gatpp}. The empirical results show that \dmn \, outperforms \magat\, in both simple and complex  maps, resulting in 5\% and 30\% increase in action selection accuracy respectively. We present results to show the effect of varying the embedding size. We also demonstrate how our model performs in the case of a noisy sensor model. 





\section{Related Works}
\label{rel_work}

Classical methods for navigation include sampling based techniques, such as Rapidly Exploring Random Tree~\cite{rrt} and Probabilistic Roadmap Planner~\cite{prm}, and graph based algorithms, such as A*~\cite{astar} and D*~\cite{dstar}, which is a variant of A* algorithm suitable for partially known environments. Learning-based methods for navigation in an occupancy grid environment mostly use Reinforcement Learning (RL), along with a few techniques based on supervised learning. The performance of such models is greatly enhanced when the models leverage memory, i.e., when an implicit or explicit representation of the occupancy grid is stored and used to select the action to be taken. 


Several memory-enabled RL based architectures have been proposed~\cite{neuralmap, hierarchy, prmrl, complexenv} for single robot navigation. In ~\cite{neuralmap}, an explicit memory representation is used to learn how to read/write the memory according to each observation. In~\cite{hierarchy, prmrl}, a memory hierarchy is stored where the levels correspond to the occupancy grid at different resolutions. The RL based models suffer from high training time and the complexity of designing a custom reward function. 

Value Iteration Network (VIN)~\cite{tamar2016vin} is a supervised learning-based model for single-robot navigation. Given the entire occupancy grid as input, VIN emulates the value iteration process~\cite{bellman1957dp} of classical shortest path algorithms to determine the action. Since its inception, VIN has been used to solve different variants of the navigation problem. For example, in ~\cite{macnarbaz, cogmapandplan}, VIN is applied to solve single-robot navigation task with partial observability.

The above-mentioned works do not address the multi-robot navigation scenario, and also not the case where the robots share information when they are close to each other. Multi-robot navigation with communication is studied in~\cite{li2020gnnpp, li2021gatpp}, but these architectures are not memory-enabled. Here, each robot constructs an implicit representation of the environment based on the observation and the received messages of only the current time-step. Hu et al.~\cite{sadler2021loop} studies the flocking problem in relevant context and uses Gated Recurrent Unit (GRU) to maintain an implicit representation of the environment over a prolonged period. All these works employ an MLP on top of the implicit memory to select the action. But an MLP is less powerful than a VIN as action selector. Consequently these works suffer from poor performance as we show in this paper. 

In our proposed architecture, we use a memory maintenance module which maintains an encoded representation of the occupancy grid, which can be decoded to reconstruct the explicit occupancy grid. We store an encoding of the grid, instead of an uncompressed grid, to decrease communication overhead. We use a convolutional autoencoder~\cite{cae} as the encoder-decoder module, which is a type of neural network used for data compression. We show that an encoding that is significantly smaller than the full map is still sufficient for accurate map maintenance and navigation.

\section{Problem Formulation}
\label{formulation}

We consider a 2D grid-world environment for multi-robot path planning. The grid-world is represented by an $X \times Y$ occupancy grid $\mathcal{O}$. Each cell of the occupancy grid is either free or occupied by an obstacle. Let $R = \{ r_1, r_2, \ldots r_N\}$ be a set of $N$ robots, each of which has a pair of start and goal positions. Robot $r_i$ starts at cell $s_i$ and is tasked to find an obstacle free path to goal cell $g_i$. Each robot can localize itself within the grid-world, i.e., it knows which cell it is in.

\textbf{Observation Model}: At each time-step, each robot makes a partial observation of the environment. An observation is limited by a receptive field spanning $Z$ cells in each direction, i.e., a $2Z+1 \times 2Z+1$ square grid centered at the robot. However we represent an observation by an $X \times Y$ grid where cells outside the receptive field are considered to be free as shown in Figure~\ref{fig_arch}. The observation made by the $i^{th}$ robot at time-step $t$ is denoted by $o_i^t$. Our proposed architecture works both for noise-free and noisy observation models. We implement a noisy observation model by flipping the occupancy status of each grid-cell within the receptive field with a probability $p$. Results for a noisy observation model is presented in Section~\ref{noisy_sensor}.

\textbf{Communication Model}: A pair of robots can exchange messages with each other if the distance between them is within the communication range $c$. Communication between a pair of neighboring robots happens instantly, and is not obstructed by obstacles. We assume a disk model for simplifying the setup, but it can be replaced by other communication model. 

\textbf{Action Transition Model}: The action taken by the $i^{th}$ robot at time-step $t$ is denoted by $a_i^t$. An action moves a robot to one of the eight adjacent cells. We assume that action transitions are deterministic and do not depend on the observations. But our proposed model can also handle a stochastic action transition model, because VINs are capable of learning action transition probabilities. 

We model the multi-robot path planning problem as a sequential decision making problem. At each time-step $t$, each robot $r_i$ makes an observation $o_i^t$ using on-board sensors, exchanges messages with other robots located within its communication range, and takes an action $a_i^t$ towards $g_i$ informed by the observation and the received messages.


Our objective is to propose a learning-based architecture to plan minimum time paths for the robots to reach their respective goal positions. Since the robots start off without knowing the map, building a map and planing their paths based on the map can be helpful. Furthermore, as the robots exchange their maps when they are close to each other, we store a compact representation of the map to reduce the communication overhead between the robots. 

\section{Architecture}
\label{architecture}

\subsection{Overview}

In our problem, the robots do not know the occupancy grid $\mathcal{O}$ initially. They learn $\mathcal{O}$ by making observations and receiving messages from other robots. A robot plans a path to its goal based on its belief about $\mathcal{O}$. Each robot independently maintains its belief of $\mathcal{O}$ using an embedding. The embedding of $r_i$ at time-step $t$ is an $H \times 1$ vector denoted by $m_i^t$. The embeddings get updated as robots make observations and receive messages from neighbors. Maintaining the embeddings enables the robots to retain long-term memory, and thus, plan better paths. 

At the beginning of time-step $t$ each robot makes a partial observation $o_i^t$ using on-board sensors, sends its embedding from previous time-step $m_i^{t-1}$ to its neighboring robots, and receives embeddings from its neighbors. Based on the sensor observation and the received messages, a robot updates its embedding. The update happens in two phases. In the first phase, the robot's embedding from the previous time-step is combined with the current observation to form an intermediate embedding $^*h_i^t$, which also has the size ($H \times 1$). In the second phase the messages received from the neighboring robots are merged with the intermediate embedding to compute the final embedding $m_i^t$. After the final embedding is computed, it is decoded to reconstruct the $X \times Y$ sized belief map, which we denote by $b_i^t$. $b_i^t$ represents the belief of robot $i$ about the occupancy status of the environment at time-step $t$. 

Given the belief map and the current and goal locations, the VIN module is used to select the action. The VIN module consists of a value iterator block which is executed $K$ times to compute the Q-values for all (state, action) pairs, and an attention block to select the optimal action using the Q-values. After the action is executed, $m_i^t$ is sent to the neighboring robots at the beginning of the next time-step. 

We divide the architecture into two modules: (i) Memory Maintenance (MM) module, which maintains the embeddings according to the observations and messages received from neighboring robots, and (ii) Value Iteration Network (VIN) module which selects the action based on the belief map.  The two modules are described in details below. Please refer to Figure~\ref{fig_arch} for a better understanding of the architecture.

\subsection{Memory Maintenance Module} 

This module takes as input the ego robot $r_i$'s embedding of the previous time-step $m_i^{t-1}$, and incorporates the observation of the current time-step and the messages received from the neighboring robots to form the new embedding $m_i^t$. Note that, the message received from a neighboring robot $r_j$ is $r_j$'s embedding of the previous time-step, $m_j^{t-1}$.

The MM module contains three types of neural network blocks: an encoder, a decoder, and an aggregator block. The encoder compresses an occupancy grid of size $X \times Y$ to an embedding vector of size $H \times 1$. The decoder reconstructs an embedding vector of size $H \times 1$ to from an occupancy grid of size $X \times Y$. The aggregator merges two embeddings of size $H \times 1$ to form another embedding of the same size. The encoder and decoder blocks can be implemented either as an MLP or a convolutional autoencoder. The aggregator block is implemented as an MLP.

Now we describe the workflow of the MM module. We represent free and occupied grid-cells by 0 and 1 respectively. The observations $o_i^t$ are represented by $X \times Y$ vectors, because although the robots make partial observations, we set the values of the unobserved grid cells to 0 as shown in Figure~\ref{fig_arch}. The observation is passed through an encoder to form a compressed representation $\hat{h}_i^t$ of size $H \times 1$. $\hat{h}_i^t$ is combined with the embedding from previous time-step $m_i^{t-1}$ to form the intermediate embedding $^*h_i^t$ of size $H \times 1$ using an aggregator, which we call observation aggregator. Then the messages received from neighboring robots are merged with the intermediate embedding one by one to form the final embedding $m_i^t$. This aggregator is called message aggregator. At last, the final embedding is decompressed using the decoder to obtain the belief map $b_i^t$.

\subsection{Value Iteration Module}  
This module takes as input $r_i$'s belief map $b_i^t$, the current location $(x_i^t, y_i^t)$, and goal location $(g_i.x, g_i.y)$, and selects one out of eight actions, one per adjacent cell.


First we provide a brief overview of the value iteration algorithm. Value iteration is a standard algorithm to solve a Markov Decision Process (MDP)~\cite{mdp}. MDPs provide a standard way to model sequential decision making problems, including navigation. In an MDP, the \textit{value} of a state $s \in \mathcal{S}$ under policy $\pi$, denoted by $V^\pi (s)$, is the expected discounted sum of rewards starting from $s$ and
taking actions according to policy $\pi$. The \textit{Q-value} of a (station, action) pair $(s, a)$, where $s \in \mathcal{S}$ and $a \in \mathcal{A}$, under policy $\pi$ is the expected discounted sum of rewards taking action $a$ starting from $s$ and executing policy $\pi$ thereafter. An optimal policy $\pi^*$ corresponds to an optimal value function $V^* (s) = \text{max}_\pi V^\pi (s), \forall s \in S$. $\pi^*$ and $V^*$ can be computed using the value iteration algorithm as follows:
\begin{equation}
\label{viequation1}
V_{n+1}(s) = \text{max}_a\,\,Q_{n}(s, a) \,\,\,\,\, \forall s \in \mathcal{S}
\end{equation}
\begin{equation}
\label{viequation2}
Q_n(s, a) = R(s, a) + \lambda \sum_{s'} P(s'|s, a)\,\, V_n(s')
\end{equation}

As $n$ approaches $\infty$, $V$ converges to $V^*$. Then the optimal policy is given by $\pi^* = \text{arg max}_a Q_{\infty}(s, a)$.

In the case of our grid-world navigation problem, the states and actions correspond to the grid-cells and movement to neighbor cells. The reward function $R$ provides high reward at the goal cell, negative reward at occupied cells, and 0 elsewhere. The transition model encodes deterministic movement on the grid and does not depend on the observation. 

Observe that in our problem, the value function has a local connectivity structure, because the value of a cell depends only on the values of the adjacent cells. Thus each iteration of the value iteration algorithm is analogous to passing the previous values ($V$) and rewards ($R$) through a convolution layer (Equation~\ref{viequation2}) and a max-pooling layer (Equation~\ref{viequation1}). Here, each channel in the convolution layer corresponds to the Q-value for a specific action. Executing the value iterator block $K$ times is equivalent to performing $K$ iterations of the value iteration loop. Thus the value iterator block outputs an $X \times Y \times A$ array of Q-values. Here $A$ is the number of actions, which equals 8. The Q-values are queried at the current location of the robot using an attention block to determine the action to be taken. The attention block is implemented as a single layer of neurons followed by a softmax layer which produces a probability distribution over which action to take.

\subsection{Training}
We follow Centralized Training and Decentralized Execution (CTDE) approach to train and deploy \dmn. We train the two modules separately and in a supervised manner. 

\begin{figure}[ht]
\centering
\includegraphics[width=1.0\linewidth]{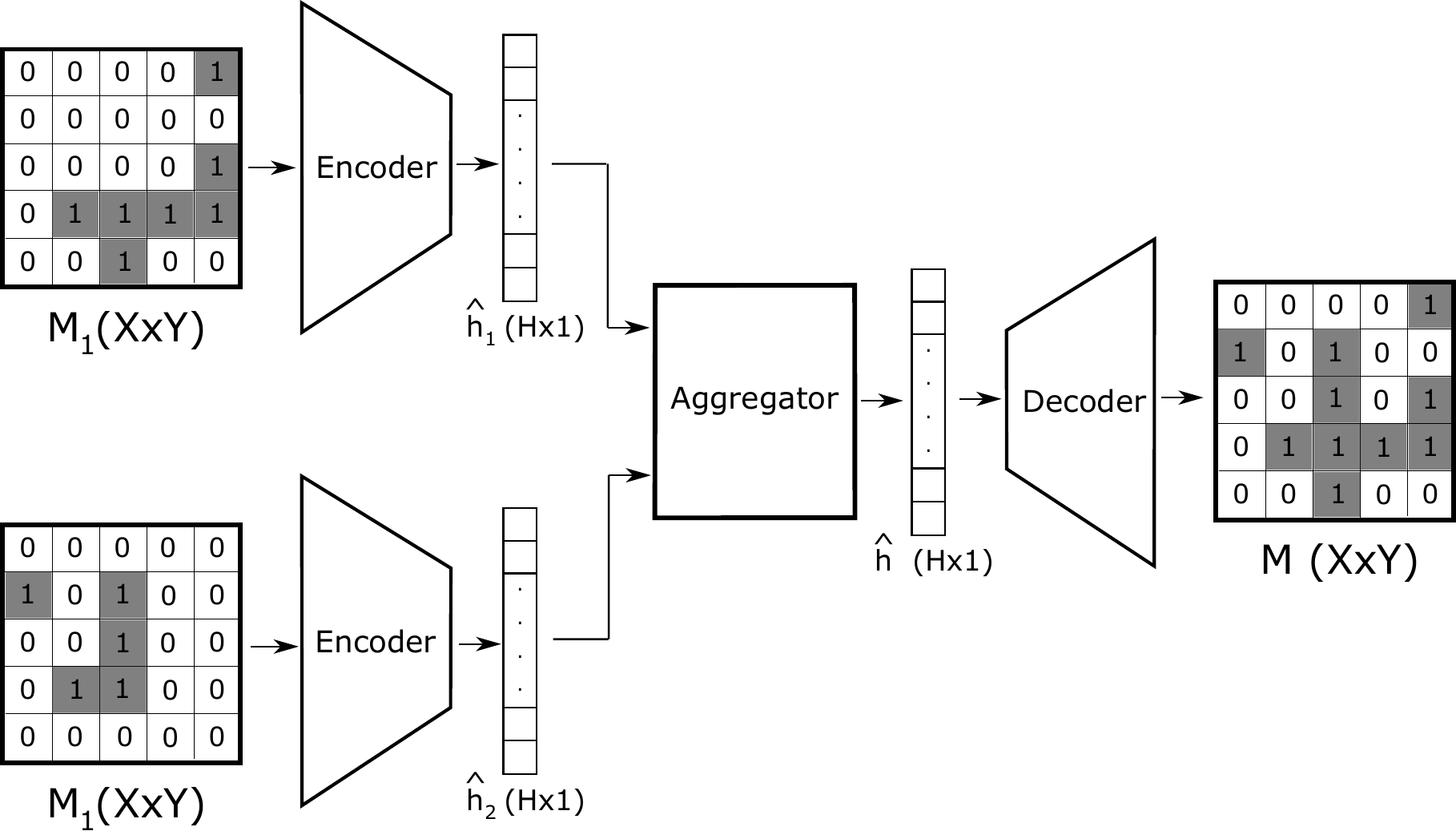}
\caption{Training the encoder, decoder, and aggregator of the MM module.}
\label{fig_train}
\end{figure}

In the MM module the message aggregator and the observation aggregator are functionally identical. They both take as input two encoded belief maps and produce the encoding of the OR of the two input belief maps. Both aggregators perform the OR operation in the implicit embedding space instead of the original grid space. In light of that, we train the MM module using the setup described in Figure~\ref{fig_train}. We generate binary maps $M_1$ and $M_2$ of size $X \times Y$ and compute $M = M_1 \,\, \text{OR} \,\, M_2$. Each tuple $(M_1, M_2, M)$ makes one training sample. We train the encoder, decoder, and aggregator by minimizing Binary Cross Entropy (BCE) loss.

To train the VIN module, we use an expert algorithm that selects the optimal action given the occupancy grid, the current location, and the goal location. The expert algorithm uses Breadth First Search~\cite{cormen} to select the optimal action. Using the expert algorithm we generate a training set consisting of (occupancy grid, current location, goal location, action) tuples. We use the training set to train the VIN module in a supervised manner by minimizing the cross entropy loss defined over the parameters of the VIN.


    
    
    
    


\section{Experiments}
\label{experiments}

In this section, first we describe the experimental setup (Section~\ref{exp_setup}). Next we provide a qualitative example to demonstrate how our proposed model works (Section~\ref{qual_example}). Finally we present the empirical results in Section~\ref{emp_res}.

\subsection{Experimental Setup}
\label{exp_setup}

\subsubsection{Compared Algorithms}

We compare our \dmn \; model with the \magat \, model proposed by Li et al.~\cite{li2021gatpp}, which is described in Section~\ref{rel_work}. We also compare our work with an Oracle solution, \oracle. \oracle \, does not compress the belief maps before observation processing and message passing. Hence, \oracle \, does not incur any error during observation and message aggregation. However, \oracle \, requires sharing the full belief maps which is significantly larger than the compact embeddings shared by \dmn \, and \magat. \oracle \, uses the true belief map to compute the optimal path using Breadth First Search~\cite{cormen}. \oracle \, deals with partial observability by assuming that the unobserved grid-cells are free.

\subsubsection{Evaluation Metric}

We evaluate the performance of \dmn\, by using two evaluation metrics: (i) Action Selection Accuracy (ASA), and (ii) Success weighted by inverse Path Length (SPL). ASA gives the percentage of time-steps a correct action is selected by a model. If there are multiple correct actions, selecting any one of them suffices. On the other hand, SPL~\cite{spl} evaluates the performance of a model at the path level. If a robot is able to reach the goal cell within thrice the optimal number of time-steps, we define it as \textit{success}. Similar definition of episodic success was used by Li et al. in~\cite{li2020gnnpp, li2021gatpp}. If a robot moves to an occluded cell or goes into a cyclic deadlock, it fails to succeed. Given this definition of success, SPL compares the length of a robot’s path with the shortest path and weighs this ratio by the success rate. Formally, given $T$ paths,
$$\text{SPL} = \frac{1}{T}\sum_{i  = i}^{T}S_i\frac{L_i}{\text{max}(P_i, L_i)}$$

Here, $S_i$ is 1 if the robot is successful in the $i^{th}$ path, and 0 otherwise. $P_i$ is the length of the robot’s path, and $L_i$ is the length of the shortest path. SPL scores range from 0 to 1, where a score of 1 represents an optimal agent.

\begin{figure}[ht]
\centering
\includegraphics[width=0.90\linewidth]{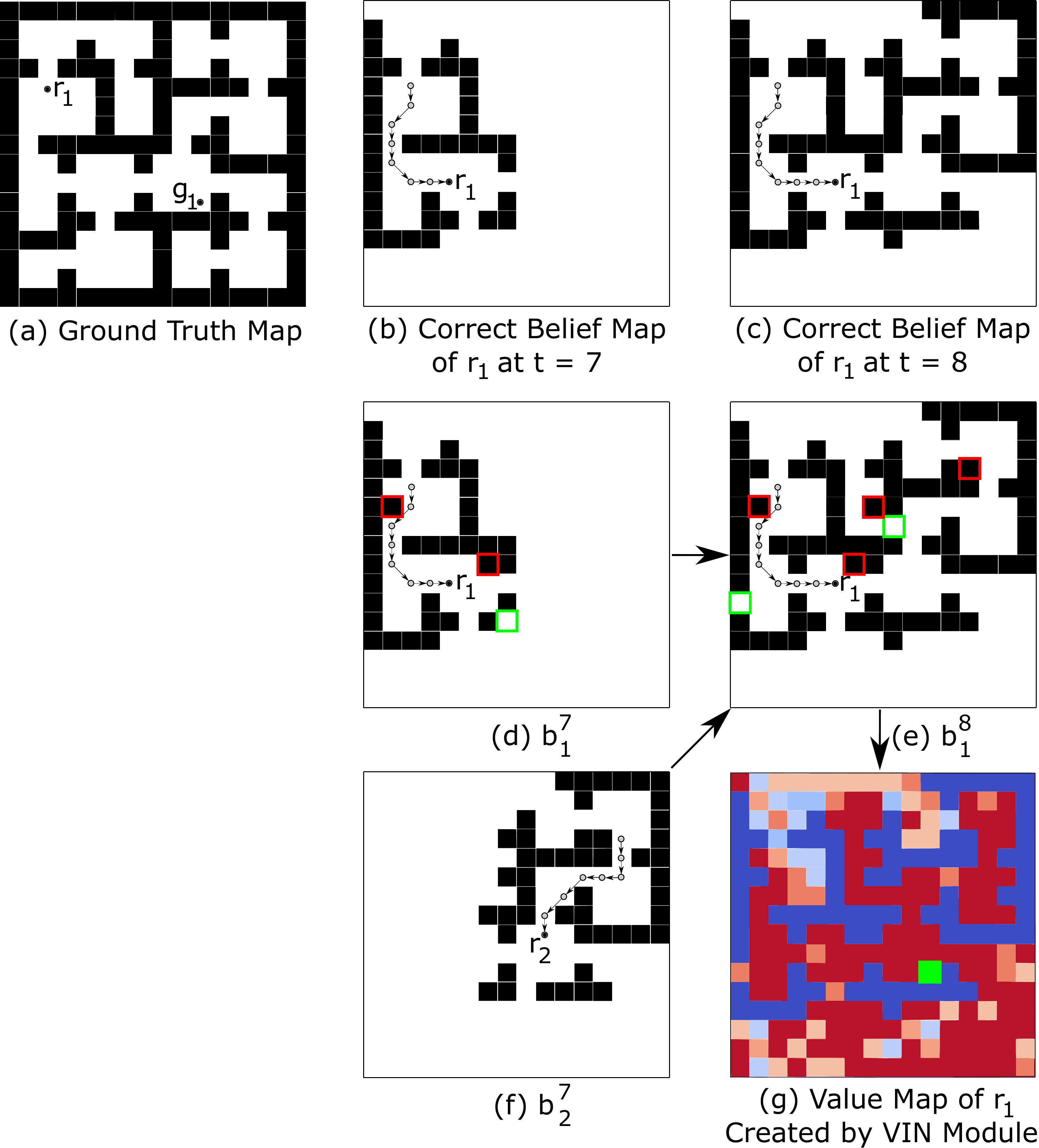}
\caption{(a) shows the occupancy grid. Robot 1 is marked as $r_1$, and $g_1$ is the goal location of $r_1$. (b) and (c) show the correct belief map of $r_1$ after 7 and 8 time-steps respectively. This would be the case if no error was incurred during observation and message aggregation. (d) shows the actual belief map of $r_1$ after 7 time-steps. False-positives and false-negatives are marked using red and green boundaries respectively. During the $8^{th}$ time-step, $r_1$ receives a message from $r_2$, which is the encoded version of $r_2$'s belief map after 7 time-steps as shown in (f). After making an observation and receiving a message from $r_2$, $r_1$'s actual belief map after time-step 8 is shown in (e). The VIN module takes $b_1^8$ and the goal location (marked by green square) as input to compute the value map as shown in (g), which is used to select the correct action for $r_1$. In (g), for each cell, we show the softmax probability of the action with the maximum Q-value, which is a measure of the model's confidence about an action. Here red and blue correspond to high and low confidence respectively. Observe that grid-cells located close to the goal cell has high confidence and vice versa.}
\label{fig_example}
\end{figure}

\begin{figure}[ht]
\centering
\includegraphics[width=0.60\linewidth]{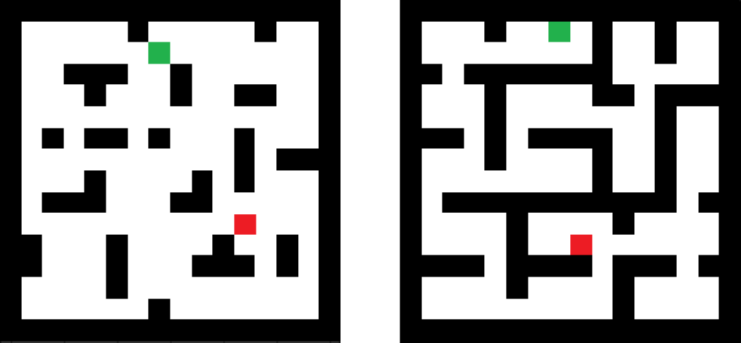}
\caption{Instances from \textsc{Simple} (left) and \textsc{Complex} (right) datasets. Green and red blocks represent source and goal cells respectively.}
\label{fig_2dataset}
\end{figure}

\begin{table*}[]
\centering
\begin{tabular}{ccccccccc}
\toprule
\multirow{3}{*}{\begin{tabular}[c]{@{}c@{}}Environment \\ Size\end{tabular}} &
  \multicolumn{4}{c}{\textsc{Simple} Dataset} &
  \multicolumn{4}{c}{\textsc{Complex} Dataset} \\[0.1cm]
 &
  \multicolumn{2}{c}{\dmn} &
  \multicolumn{2}{c}{\magat} &
  \multicolumn{2}{c}{\dmn} &
  \multicolumn{2}{c}{\magat} \\[0.1cm]
      & ASA (\%) & SPL  & ASA (\%) & SPL  & ASA (\%) & SPL  & ASA (\%) & SPL  \\
      \midrule
12x12 & 97  & 0.97 & 94  & 0.85 & 96  & 0.96 & 73  & 0.18 \\
16x16 & 96  & 0.92 & 90  & 0.71 & 93  & 0.83 & 71  & 0.16 \\
20x20 & 93  & 0.83 & 88  & 0.67 & 91  & 0.73 & 67  & 0.11 \\
24x24 & 91  & 0.74 & 86  & 0.64 & 82  & 0.57 & 61  & 0.05
\\
\bottomrule
\end{tabular}
\caption{Performance of the VIN module in single robot navigation task under full observability}
\label{vin_res}
\end{table*}

\subsubsection{Dataset}
We use two datasets to evaluate the performance of our proposed architecture, namely \textsc{Simple} and \textsc{Complex} dataset. In the \textsc{Simple} dataset, the environment contains tetris block shaped obstacles, and the (source, goal) pairs are chosen such that the distance between each pair is less than 1.5 times their unobstructed Manhattan distance. The \textsc{Simple} dataset is similar to the dataset used by Li et al. in~\cite{li2020gnnpp, li2021gatpp}. However, in their dataset, the percentage of grid-cells that are occluded is 10\%, whereas in our \textsc{Simple} dataset, roughly 20\% of the grid is occluded. In the case of \textsc{Complex} dataset, we generate random floorplan-like occupancy grids and the distance between each (source, goal) pair is at least twice their unobstructed Manhattan distance. The percentages of occupied grid cells in  \textsc{Complex} dataset is roughly 35\%. Please refer to Figure~\ref{fig_2dataset} to see samples of both datasets. The dataset size varies from 300K to 1.5M depending on the environment size with a 5:1 train-test split.

\subsubsection{Platform}
The experiments are performed using a Core i9-12900H processor and an Nvidia GeForce RTX 3070 Ti GPU with 16GB RAM running Windows 11.

\subsection{A Qualitative Example}
\label{qual_example}

In this section, we demonstrate how \dmn\, works with an illustrative example shown in Figure~\ref{fig_example}. The example shows how robot $r_1$'s belief map after 7 time-steps gets updated as it receives a message from robot $r_2$ during the $8^{th}$ time-step. In this example, the receptive field of $r_1$ is $7 \times 7$. Note that, the belief map of robot $r_i$ after time-step $t$ is denoted by $b_i^t$. 

\vspace{-0.1cm}

\subsection{Empirical Results}
\label{emp_res}

First we report the performance of the two modules independently (Section~\ref{perf_modules}). Then we cascade the two modules to solve the multi-robot navigation task and present the results in Section~\ref{mrn_task}. Finally we report the performance of \dmn \, for a noisy sensor model (Section~\ref{noisy_sensor}) and for a navigation task with multiple goals (Section~\ref{multi_goal}).

\subsubsection{Performance of Individual Modules}
\label{perf_modules}

First we present the performance of the MM module for the setup described in Figure~\ref{fig_train}. As dataset, we use random subsets or windows from the \textsc{Complex} dataset. In this experiment, we vary the environment size and the size of the embedding, $H$, and report the percentage of correctly reconstructed grid cells in Table~\ref{mm_res}. The results show that our MM module has reconstruction accuracy of more than 90\% when a $24 \times 24$ map is compressed to a $16 \times 1$ vector, which gives a compression ratio is 36. The results show expected behavior as reconstruction accuracy decreases if the environment size is increased or the value of $H$ is lowered, and vice versa.

Now we present the performance of the VIN module in single robot navigation task under full observability. We compare our architecture with \magat~\cite{li2021gatpp} using both \textsc{Simple} and \textsc{Complex} datasets of size $12 \times 12$ to $24 \times 24$ and report the ASA and SPL values in Table~\ref{vin_res}. The results show that \dmn\, outperforms \magat\, for both datasets, but the margin of difference is significantly higher in the case of \textsc{Complex} datasets. For example, the SPL value of \dmn \, is at least 5 times higher than \magat \, for \textsc{Complex} dataset. This corroborates the fact that VIN is better at solving navigation task in complex environments than existing MLP based models, justifying our choice for using VIN as the planner. Next, we will evaluate how well the MM module works in conjunction with VIN.

\begin{table}[]
\centering
\begin{tabular}{ccccc}
\toprule
Environment Size & H=16 & H=32 & H=64 & H=128 \\
\midrule
12x12            & 97.8 & 99.2 & 99.8 & 99.9  \\
16x16            & 96.9 & 98.8 & 99.6 & 99.9  \\
20x20            & 93.3 & 97.7 & 99.2 & 99.8  \\
24x24            & 90.1 & 95.8 & 98.4 & 99.4  \\
\bottomrule
\end{tabular}
\caption{Performance of the MM module}
\label{mm_res}
\end{table}

\begin{figure*}[]
\centering
\includegraphics[width=0.8\linewidth]{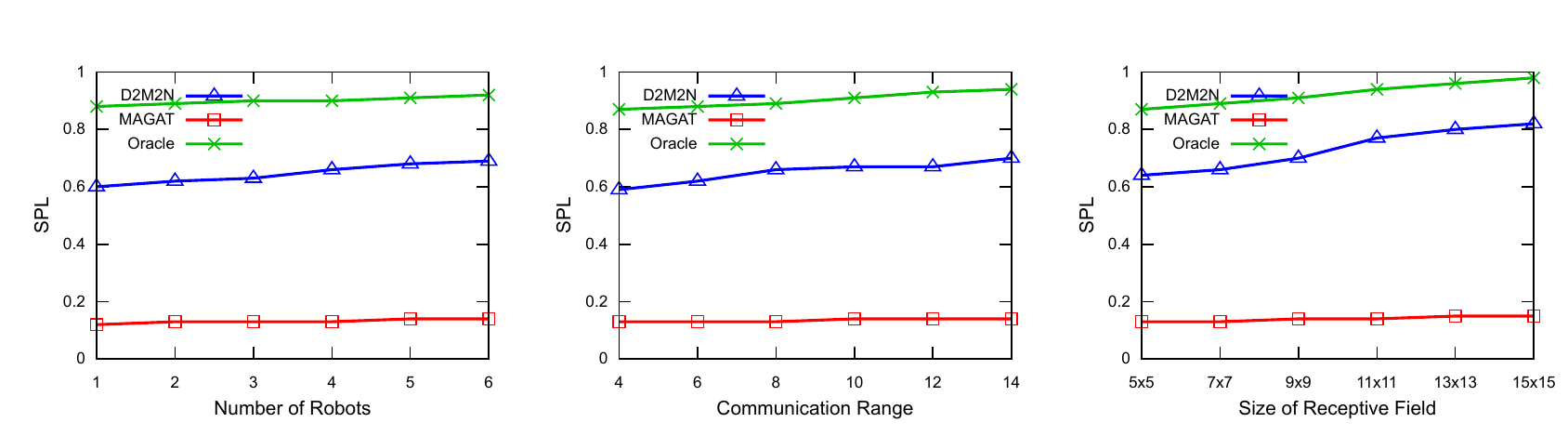}
\caption{Effect of varying the number of robots (left), communication range (middle), and the size of the receptive field (right).}
\label{fig_3exp}
\end{figure*}

\subsubsection{Performance in Multi-Robot Navigation Task}
\label{mrn_task}

We cascade the MM and VIN modules to solve the multi-robot navigation task under partial observability. We incorporate partial observability by shrinking the receptive field of the robots. As a result, a robot cannot observe the entire occupancy grid, instead it observes a square shaped region centered at its location (see Figure~\ref{fig_arch}). In all the experiments of this section, we use a $16 \times 16$ environment and an embedding of size $H=32$. We vary the number of robots from 1 to 6, the communication range of the robots from 4 to 14 (Manhattan distance), and the size of the receptive field from $5 \times 5$ to $15 \times 15$. The default values of the number of robots, communication range, and receptive field size are 4, 8, and $7 \times 7$, respectively. We use the same value for the parameters in subsequent experiments, unless mentioned otherwise.

We compare \dmn\, with \magat\, and \oracle\, using \textsc{Complex} dataset and report the SPL values in Figure~\ref{fig_3exp}. The results show that the SPL value of \dmn \, is at least 5 times higher than \magat. \dmn \, performs better than \magat\, because it stores memory of past observations which \magat \, does not, and uses a VIN to select actions instead of an MLP. On the other hand, the SPL value of \oracle \, is roughly 0.2 higher than \dmn. \oracle\, outperforms \dmn\, because it uses the full map as messages while \dmn\, shares compressed embeddings, and \oracle\, uses an optimal path-planning algorithm while \dmn\, uses a VIN. The performance of our model improves as the number of robots, communication range, or receptive field size increases. Because if the number of robots or the communication range increases, inter-robot message passing increases, which helps the robots learn about unknown regions in the environment. Increasing the receptive field widens observability, and hence leads to better performance.

\begin{table}[]
\centering
\begin{tabular}{ccc}
\toprule
Noise (p) & \dmn \, (SPL) & \magat \, (SPL) \\
\midrule
0         & 0.66      & 0.13            \\
0.005     & 0.63      & 0.13            \\
0.01      & 0.59      & 0.12            \\
0.02      & 0.53      & 0.11            \\
0.05      & 0.42      & 0.10            \\
\bottomrule
\end{tabular}
\caption{Performance for a noisy sensor model.}
\label{exp_sensor}
\end{table}

\begin{table}[]
\centering
\begin{tabular}{ccc}
\toprule
Goal Index & \dmn \, (SPL) & \magat \, (SPL) \\
\midrule
1          & 0.60      & 0.13            \\
2          & 0.66      & 0.13            \\
3          & 0.70      & 0.13            \\
4          & 0.74      & 0.13            \\
5          & 0.78      & 0.13            \\
\bottomrule
\end{tabular}
\caption{Performance for multiple goals}
\label{exp_mult}
\end{table}

\subsubsection{Noisy Sensor Model}
\label{noisy_sensor}

In the above experiments, we assume that each robot correctly captures the occupancy status of the environment within it's receptive field. In this experiment, we relax this assumption and evaluate how our model performs in the case of a noisy sensor model. To simulate a noisy sensor model, in each observation $o_i^t$, we randomly flip the status (free or occupied) of grid cells inside the receptive field with a probability $p$. We vary $p$ from 0 (which corresponds to no noise) to 0.05. The results (Table~\ref{exp_sensor}) show that even with 5\% noise, \dmn \, performs four times better than \magat \, under the SPL metric.

\subsubsection{Multi-Goal Navigation}
\label{multi_goal}

In the final experiment, we provide each robot with multiple goals, instead of one goal. The task is to reach the first goal, then go the second goal, and so on. The results in Table~\ref{exp_mult} show that performance improves as the goal index increases. This is because as a robot spends more time navigating the environment, it learns more about the occupancy status, hence becomes able to plan better paths later compared to the initial phase. Note from Table~\ref{vin_res} that the SPL value of \dmn \, for full observability in $16 \times 16$ environment is 0.83. Thus the SPL value of \dmn \, approaches 0.83 as the index of the goal increases.

\section{CONCLUSIONS}

In this work, we have proposed \dmn, a decentralized and differential model for multi-robot mapping and navigation, which, unlike previous works, includes a memory unit to remember past observations and uses a VIN, instead of an MLP, for action selection. We have conducted experiments to show that \dmn \, significantly outperforms the state-of-the-art~\cite{li2021gatpp} in complex navigation task. In future, we intend to extend \dmn \, to handle inter-robot collision and perform experiments on larger realistic environments.






\bibliographystyle{IEEEtran} 

@inproceedings{cogmapandplan,
  title={Cognitive mapping and planning for visual navigation},
  author={Gupta, Saurabh and Davidson, James and Levine, Sergey and Sukthankar, Rahul and Malik, Jitendra},
  booktitle={Proceedings of the IEEE conference on computer vision and pattern recognition},
  pages={2616--2625},
  year={2017}
};

@article{neuralmap,
  title={Neural map: Structured memory for deep reinforcement learning},
  author={Parisotto, Emilio and Salakhutdinov, Ruslan},
  journal={arXiv preprint arXiv:1702.08360},
  year={2017}
};

@article{complexenv,
  title={Learning to navigate in complex environments},
  author={Mirowski, Piotr and Pascanu, Razvan and Viola, Fabio and Soyer, Hubert and Ballard, Andrew J and Banino, Andrea and Denil, Misha and Goroshin, Ross and Sifre, Laurent and Kavukcuoglu, Koray and others},
  journal={arXiv preprint arXiv:1611.03673},
  year={2016}
};

@article{prmrl,
  title={Long-range indoor navigation with prm-rl},
  author={Francis, Anthony and Faust, Aleksandra and Chiang, Hao-Tien Lewis and Hsu, Jasmine and Kew, J Chase and Fiser, Marek and Lee, Tsang-Wei Edward},
  journal={IEEE Transactions on Robotics},
  volume={36},
  number={4},
  pages={1115--1134},
  year={2020},
  publisher={IEEE}
};

@inproceedings{hierarchy,
  title={Hierarchical representations and explicit memory: Learning effective navigation policies on 3D scene graphs using graph neural networks},
  author={Ravichandran, Zachary and Peng, Lisa and Hughes, Nathan and Griffith, J Daniel and Carlone, Luca},
  booktitle={2022 International Conference on Robotics and Automation (ICRA)},
  pages={9272--9279},
  year={2022},
  organization={IEEE}
};


@article{macnarbaz,
  title={Memory augmented control networks},
  author={Khan, Arbaaz and Zhang, Clark and Atanasov, Nikolay and Karydis, Konstantinos and Kumar, Vijay and Lee, Daniel D},
  journal={arXiv preprint arXiv:1709.05706},
  year={2017}
};

@incollection{dstar,
  title={Optimal and efficient path planning for partially known environments},
  author={Stentz, Anthony},
  booktitle={Intelligent unmanned ground vehicles},
  pages={203--220},
  year={1997},
  publisher={Springer}
};

@article{rrt,
  title={Rapidly-exploring random trees : a new tool for path planning},
  author={Steven M. LaValle},
  journal={The annual research report},
  year={1998}
};

@article{prm,
  title={Probabilistic roadmaps for path planning in high-dimensional configuration spaces},
  author={Kavraki, Lydia E and Svestka, Petr and Latombe, J-C and Overmars, Mark H},
  journal={IEEE transactions on Robotics and Automation},
  volume={12},
  number={4},
  pages={566--580},
  year={1996},
  publisher={IEEE}
};

@ARTICLE{astar,
  author={Hart, Peter E. and Nilsson, Nils J. and Raphael, Bertram},
  journal={IEEE Transactions on Systems Science and Cybernetics}, 
  title={A Formal Basis for the Heuristic Determination of Minimum Cost Paths}, 
  year={1968},
  volume={4},
  number={2},
  pages={100-107},
  doi={10.1109/TSSC.1968.300136}
};

@article{drsurveyae,
  title={Auto-encoder based dimensionality reduction},
  author={Wang, Yasi and Yao, Hongxun and Zhao, Sicheng},
  journal={Neurocomputing},
  volume={184},
  pages={232--242},
  year={2016},
  publisher={Elsevier}
};

@article{drsurveycl,
  title={Dimensionality reduction: a comparative Review},
  author={Van Der Maaten, Laurens and Postma, Eric and Van den Herik, Jaap and others},
  journal={Journal of Machine Learning Research},
  volume={10},
  number={66-71},
  pages={13},
  year={2009}
};

@article{spl,
  title={On evaluation of embodied navigation agents},
  author={Anderson, Peter and Chang, Angel and Chaplot, Devendra Singh and Dosovitskiy, Alexey and Gupta, Saurabh and Koltun, Vladlen and Kosecka, Jana and Malik, Jitendra and Mottaghi, Roozbeh and Savva, Manolis and others},
  journal={arXiv preprint arXiv:1807.06757},
  year={2018}
};

@article{mdp,
  title={Dynamic programming and markov processes.},
  author={Howard, Ronald A},
  year={1960},
  publisher={John Wiley}
};

@book{cormen,
  author    = {Thomas H. Cormen and
               Charles E. Leiserson and
               Ronald L. Rivest and
               Clifford Stein},
  title     = {Introduction to Algorithms, 3rd Edition},
  publisher = {{MIT} Press},
  year      = {2009},
  url       = {http://mitpress.mit.edu/books/introduction-algorithms},
  isbn      = {978-0-262-03384-8},
  timestamp = {Mon, 17 Aug 2020 01:00:00 +0200},
  biburl    = {https://dblp.org/rec/books/daglib/0023376.bib},
  bibsource = {dblp computer science bibliography, https://dblp.org}
};


@article{drwithnn,
  title={Reducing the dimensionality of data with neural networks},
  author={Hinton, Geoffrey E and Salakhutdinov, Ruslan R},
  journal={Science},
  volume={313},
  number={5786},
  pages={504--507},
  year={2006},
  publisher={American Association for the Advancement of Science}
};

@inproceedings{flocking1,
  title={Learning decentralized controllers for robot swarms with graph neural networks},
  author={Tolstaya, Ekaterina and Gama, Fernando and Paulos, James and Pappas, George and Kumar, Vijay and Ribeiro, Alejandro},
  booktitle={Conference on robot learning},
  pages={671--682},
  year={2020},
  organization={PMLR}
};

@inproceedings{exploration1,
  title={Autonomous exploration under uncertainty via deep reinforcement learning on graphs},
  author={Chen, Fanfei and Martin, John D and Huang, Yewei and Wang, Jinkun and Englot, Brendan},
  booktitle={2020 IEEE/RSJ International Conference on Intelligent Robots and Systems (IROS)},
  pages={6140--6147},
  year={2020},
  organization={IEEE}
};

@inproceedings{coverage1,
  title={Multi-robot coverage and exploration using spatial graph neural networks},
  author={Tolstaya, Ekaterina and Paulos, James and Kumar, Vijay and Ribeiro, Alejandro},
  booktitle={2021 IEEE/RSJ International Conference on Intelligent Robots and Systems (IROS)},
  pages={8944--8950},
  year={2021},
  organization={IEEE}
};

@article{flocking2,
  title={Learning vision-based flight in drone swarms by imitation},
  author={Schilling, Fabian and Lecoeur, Julien and Schiano, Fabrizio and Floreano, Dario},
  journal={IEEE Robotics and Automation Letters},
  volume={4},
  number={4},
  pages={4523--4530},
  year={2019},
  publisher={IEEE}
};

@article{pca,
  title={Principal component analysis},
  author={Wold, Svante and Esbensen, Kim and Geladi, Paul},
  journal={Chemometrics and intelligent laboratory systems},
  volume={2},
  number={1-3},
  pages={37--52},
  year={1987},
  publisher={Elsevier}
};

@inproceedings{cae,
  title={Stacked convolutional auto-encoders for hierarchical feature extraction},
  author={Masci, Jonathan and Meier, Ueli and Cire{\c{s}}an, Dan and Schmidhuber, J{\"u}rgen},
  booktitle={International conference on artificial neural networks},
  pages={52--59},
  year={2011},
  organization={Springer}
};



@article{li2020gnnpp,
  author    = "Q. Li and F. Gama and A. Ribeiro and A. Prorok",
  title     = "Graph Neural Networks for Decentralized Multi-Robot Path Planning",
  journal   = "2020 IEEE/RSJ International Conference on Intelligent Robots and Systems (IROS)",
  year      = "2020",
  pages     = "11785-11792"
};

@article{li2021gatpp,
  author="Q. Li and W. Lin and Z. Lin and A. Prorok",
  journal="IEEE Robotics and Automation Letters", 
  title="Message-Aware Graph Attention Networks for Large-Scale Multi-Robot Path Planning", 
  year="2021",
  volume="6",
  number="3",
  pages="5533-5540"
};  
  
@article{sadler2021loop,
  author="T. Hu and F. Gama and T. Chen and W. Zheng and Z. Wang and A. Ribeiro and B. M. Sadler",
  journal="IEEE Transactions on Signal and Information Processing over Networks", 
  title="Scalable Perception-Action-Communication Loops With Convolutional and Graph Neural Networks", 
  year="2022",
  volume="8",
  pages="12-24"
};

@inproceedings{tamar2016vin,
 author = {Tamar, Aviv and WU, YI and Thomas, Garrett and Levine, Sergey and Abbeel, Pieter},
 booktitle = {Advances in Neural Information Processing Systems},
 editor = {D. Lee and M. Sugiyama and U. Luxburg and I. Guyon and R. Garnett},
 pages = {},
 publisher = {Curran Associates, Inc.},
 title = {Value Iteration Networks},
 volume = {29},
 year = {2016}
};

@book{bellman1957dp,
  author        = "Richard Bellman",
  title         = "Dynamic Programming",
  publisher     = "Princeton University Press",
  address       = "Princeton, NJ",
  year          = "1957"
};

@book{lda,
  title={Pattern classification},
  author={Hart, Peter E and Stork, David G and Duda, Richard O},
  year={2000},
  publisher={Wiley Hoboken}
};



\begin{thebibliography}{10}
\providecommand{\url}[1]{#1}
\csname url@rmstyle\endcsname
\providecommand{\newblock}{\relax}
\providecommand{\bibinfo}[2]{#2}
\providecommand\BIBentrySTDinterwordspacing{\spaceskip=0pt\relax}
\providecommand\BIBentryALTinterwordstretchfactor{4}
\providecommand\BIBentryALTinterwordspacing{\spaceskip=\fontdimen2\font plus
\BIBentryALTinterwordstretchfactor\fontdimen3\font minus
  \fontdimen4\font\relax}
\providecommand\BIBforeignlanguage[2]{{%
\expandafter\ifx\csname l@#1\endcsname\relax
\typeout{** WARNING: IEEEtran.bst: No hyphenation pattern has been}%
\typeout{** loaded for the language `#1'. Using the pattern for}%
\typeout{** the default language instead.}%
\else
\language=\csname l@#1\endcsname
\fi
#2}}

\bibitem{neuralmap}
E.~Parisotto and R.~Salakhutdinov, ``Neural map: Structured memory for deep
  reinforcement learning,'' \emph{arXiv preprint arXiv:1702.08360}, 2017.

\bibitem{hierarchy}
Z.~Ravichandran, L.~Peng, N.~Hughes, J.~D. Griffith, and L.~Carlone,
  ``Hierarchical representations and explicit memory: Learning effective
  navigation policies on 3d scene graphs using graph neural networks,'' in
  \emph{2022 International Conference on Robotics and Automation (ICRA)}.\hskip
  1em plus 0.5em minus 0.4em\relax IEEE, 2022, pp. 9272--9279.

\bibitem{prmrl}
A.~Francis, A.~Faust, H.-T.~L. Chiang, J.~Hsu, J.~C. Kew, M.~Fiser, and
  T.-W.~E. Lee, ``Long-range indoor navigation with prm-rl,'' \emph{IEEE
  Transactions on Robotics}, vol.~36, no.~4, pp. 1115--1134, 2020.

\bibitem{complexenv}
P.~Mirowski, R.~Pascanu, F.~Viola, H.~Soyer, A.~J. Ballard, A.~Banino,
  M.~Denil, R.~Goroshin, L.~Sifre, K.~Kavukcuoglu, \emph{et~al.}, ``Learning to
  navigate in complex environments,'' \emph{arXiv preprint arXiv:1611.03673},
  2016.




\bibitem{exploration1}
F.~Chen, J.~D. Martin, Y.~Huang, J.~Wang, and B.~Englot, ``Autonomous
  exploration under uncertainty via deep reinforcement learning on graphs,'' in
  \emph{2020 IEEE/RSJ International Conference on Intelligent Robots and
  Systems (IROS)}.\hskip 1em plus 0.5em minus 0.4em\relax IEEE, 2020, pp.
  6140--6147.

\bibitem{shi2021communication}
G.~Shi, I.E.~Rabban, L.~Zhou, P.~Tokekar, ``Communication-aware multi-robot coordination with submodular maximization'', \emph{2021 IEEE International Conference on Robotics and Automation (ICRA)}, pp.~8955-8961). 2021.

\bibitem{sadler2021loop}
T.~Hu, F.~Gama, T.~Chen, W.~Zheng, Z.~Wang, A.~Ribeiro, and B.~M. Sadler,
  ``Scalable perception-action-communication loops with convolutional and graph
  neural networks,'' \emph{IEEE Transactions on Signal and Information
  Processing over Networks}, vol.~8, pp. 12--24, 2022.




\bibitem{flocking1}
E.~Tolstaya, F.~Gama, J.~Paulos, G.~Pappas, V.~Kumar, and A.~Ribeiro,
  ``Learning decentralized controllers for robot swarms with graph neural
  networks,'' in \emph{Conference on robot learning}.\hskip 1em plus 0.5em
  minus 0.4em\relax PMLR, 2020, pp. 671--682.

\bibitem{flocking2}
F.~Schilling, J.~Lecoeur, F.~Schiano, and D.~Floreano, ``Learning vision-based
  flight in drone swarms by imitation,'' \emph{IEEE Robotics and Automation
  Letters}, vol.~4, no.~4, pp. 4523--4530, 2019.

\bibitem{coverage1}
E.~Tolstaya, J.~Paulos, V.~Kumar, and A.~Ribeiro, ``Multi-robot coverage and
  exploration using spatial graph neural networks,'' in \emph{2021 IEEE/RSJ
  International Conference on Intelligent Robots and Systems (IROS)}.\hskip 1em
  plus 0.5em minus 0.4em\relax IEEE, 2021, pp. 8944--8950.

\bibitem{rabban2021failure}
M.~Ishat-E-Rabban, P.~Tokekar, ``Failure-resilient coverage maximization with multiple robots'', \emph{IEEE Robotics and Automation Letters}, vol.~6(2), pp.~3894-3901, 2021.

\bibitem{rabban2019mvfs}
M.~Rabban, M.~Ali, M.~Cheema, T.~Hashem, ``The Maximum Visibility Facility Selection Query in Spatial Databases'', \emph{27th ACM Sigspatial International Conference on Advances in Geographic Information Systems}, pp.~149--158, 2019.

\bibitem{li2020gnnpp}
Q.~Li, F.~Gama, A.~Ribeiro, and A.~Prorok, ``Graph neural networks for
  decentralized multi-robot path planning,'' \emph{2020 IEEE/RSJ International
  Conference on Intelligent Robots and Systems (IROS)}, pp. 11\,785--11\,792,
  2020.

\bibitem{li2021gatpp}
Q.~Li, W.~Lin, Z.~Lin, and A.~Prorok, ``Message-aware graph attention networks
  for large-scale multi-robot path planning,'' \emph{IEEE Robotics and
  Automation Letters}, vol.~6, no.~3, pp. 5533--5540, 2021.

\bibitem{tamar2016vin}
A.~Tamar, Y.~WU, G.~Thomas, S.~Levine, and P.~Abbeel, ``Value iteration
  networks,'' in \emph{Advances in Neural Information Processing Systems},
  D.~Lee, M.~Sugiyama, U.~Luxburg, I.~Guyon, and R.~Garnett, Eds.,
  vol.~29.\hskip 1em plus 0.5em minus 0.4em\relax Curran Associates, Inc.,
  2016.

\bibitem{bellman1957dp}
R.~Bellman, \emph{Dynamic Programming}.\hskip 1em plus 0.5em minus 0.4em\relax
  Princeton, NJ: Princeton University Press, 1957.

\bibitem{rrt}
S.~M. LaValle, ``Rapidly-exploring random trees : a new tool for path
  planning,'' \emph{The annual research report}, 1998.

\bibitem{prm}
L.~E. Kavraki, P.~Svestka, J.-C. Latombe, and M.~H. Overmars, ``Probabilistic
  roadmaps for path planning in high-dimensional configuration spaces,''
  \emph{IEEE transactions on Robotics and Automation}, vol.~12, no.~4, pp.
  566--580, 1996.

\bibitem{astar}
P.~E. Hart, N.~J. Nilsson, and B.~Raphael, ``A formal basis for the heuristic
  determination of minimum cost paths,'' \emph{IEEE Transactions on Systems
  Science and Cybernetics}, vol.~4, no.~2, pp. 100--107, 1968.

\bibitem{dstar}
A.~Stentz, ``Optimal and efficient path planning for partially known
  environments,'' in \emph{Intelligent unmanned ground vehicles}.\hskip 1em
  plus 0.5em minus 0.4em\relax Springer, 1997, pp. 203--220.

\bibitem{macnarbaz}
A.~Khan, C.~Zhang, N.~Atanasov, K.~Karydis, V.~Kumar, and D.~D. Lee, ``Memory
  augmented control networks,'' \emph{arXiv preprint arXiv:1709.05706}, 2017.

\bibitem{cogmapandplan}
S.~Gupta, J.~Davidson, S.~Levine, R.~Sukthankar, and J.~Malik, ``Cognitive
  mapping and planning for visual navigation,'' in \emph{Proceedings of the
  IEEE conference on computer vision and pattern recognition}, 2017, pp.
  2616--2625.

\bibitem{cae}
J.~Masci, U.~Meier, D.~Cire{\c{s}}an, and J.~Schmidhuber, ``Stacked
  convolutional auto-encoders for hierarchical feature extraction,'' in
  \emph{International conference on artificial neural networks}.\hskip 1em plus
  0.5em minus 0.4em\relax Springer, 2011, pp. 52--59.



\bibitem{mdp}
R.~A. Howard, ``Dynamic programming and markov processes.'' 1960.

\bibitem{cormen}
\BIBentryALTinterwordspacing
T.~H. Cormen, C.~E. Leiserson, R.~L. Rivest, and C.~Stein, \emph{Introduction
  to Algorithms, 3rd Edition}.\hskip 1em plus 0.5em minus 0.4em\relax {MIT}
  Press, 2009. [Online]. Available:
  \url{http://mitpress.mit.edu/books/introduction-algorithms}
\BIBentrySTDinterwordspacing

\bibitem{spl}
P.~Anderson, A.~Chang, D.~S. Chaplot, A.~Dosovitskiy, S.~Gupta, V.~Koltun,
  J.~Kosecka, J.~Malik, R.~Mottaghi, M.~Savva, \emph{et~al.}, ``On evaluation
  of embodied navigation agents,'' \emph{arXiv preprint arXiv:1807.06757},
  2018.

\end{thebibliography}

\end{document}